\documentclass{article}

\usepackage[preprint]{neurips_2023}

\usepackage[utf8]{inputenc} 
\usepackage[T1]{fontenc}    
\usepackage{hyperref}       
\usepackage{url}            
\usepackage{booktabs}       
\usepackage{amsfonts}       
\usepackage{nicefrac}       
\usepackage{microtype}      
\usepackage{xcolor}         
\usepackage{graphicx}
\usepackage{wrapfig}
\usepackage{multirow}
\usepackage{amsmath}
\usepackage{amssymb}
\usepackage{subcaption}
\usepackage{textcomp}
\usepackage{threeparttable}
\usepackage[normalem]{ulem}
\usepackage{array}

\newcommand{\LLMEP}{\textsc{Rememberer}}
\newcommand{\Llmep}{\textsc{Rememberer}}
\newcommand{\llmep}{\textsc{Rmmbr.}}

\begin{document}

\title{Large Language Models Are Semi-Parametric Reinforcement Learning Agents}

%

\author{%
	Danyang Zhang$^1$ \And
	Lu Chen$^{1,\:\!2}$\:\!\thanks{Lu Chen is the corresponding author.} \And
	Situo Zhang$^1$ \AND
	Hongshen Xu$^1$ \And
	Zihan Zhao$^1$ \And
	Kai Yu$^{1,\:\!2}$ \and
	$^1$X-LANCE Lab, Department of Computer Science and Engineering \\
	MoE Key Lab of Artificial Intelligence, SJTU AI Institute \\
	Shanghai Jiao Tong University, Shanghai, China  \\
	$^2$Suzhou Laboratory, Suzhou, China \\
	\texttt{\{zhang-dy20,chenlusz,situozhang,xuhongshen,zhao\_mengxin,kai.yu\}@sjtu.edu.cn} \\
}

\setcounter{footnote}{1}
\setcounter{Hfootnote}{1}
\maketitle
\setcounter{footnote}{0}
\setcounter{Hfootnote}{0}

\begin{abstract}
	Inspired by the insights in cognitive science with respect to human memory
	and reasoning mechanism, a novel evolvable LLM-based (Large Language Model)
	agent framework is proposed as \LLMEP{}. By equipping the LLM with a
	long-term experience memory, \Llmep{} is capable of exploiting the
	experiences from the past episodes even for different task goals, which
	excels an LLM-based agent with fixed exemplars or equipped with a transient
	working memory.  We further introduce \textbf{R}einforcement
	\textbf{L}earning with \textbf{E}xperience \textbf{M}emory (\textbf{RLEM})
	to update the memory. Thus, the whole system can learn from the experiences
	of both success and failure, and evolve its capability without fine-tuning
	the parameters of the LLM. In this way, the proposed \Llmep{} constitutes a
	semi-parametric RL agent. Extensive experiments are conducted on two RL
	task sets to evaluate the proposed framework. The average results with
	different initialization and training sets exceed the prior SOTA by 4\% and
	2\% for the success rate on two task sets and demonstrate the superiority
	and robustness of \Llmep{}.\footnote{The codes are open-sourced at
	\url{https://github.com/OpenDFM/Rememberer}.}
\end{abstract}

\section{Introduction}

{\em Reasoning is remembering}. As declared by 
\citet{SeifertCollenM1997_ReasoningIsRemembering}, the episodic memory of the experiences from
past episodes plays a crucial role in the complex decision-making processes of human 
\citep{SuddendorfThomas2007BehavioralBrainSci_EpisodicMemory}.
By recollecting the experiences from past episodes, the human
can learn from success to repeat it and learn from failure to avoid it. Similarly, an agent should 
optimize its policy for a decision-making task with the help of reminiscence of the interaction experiences. 
In this work, we primarily investigate how to utilize large language models (LLMs) as agents and 
equip them with interaction experiences to solve sequential decision-making tasks.

Despite the impressive performance of LLMs on many natural language processing (NLP) 
tasks~\citep{JasonWei2022_CoT,TakeshiKojima2022NeurIPS_StepByStep, XuezhiWang2022_SelfConsistency,ShunyuYao2022_ReAct},
existing approaches still struggle to enable LLMs to effectively learn from interaction experiences.       
On the one hand, the most common approach for an agent to utilize the experiences is to
fine-tune the model parameters through reinforcement learning (RL). However, it requires a considerable expenditure
to deploy and fine-tune an LLM, which makes it difficult to apply task-aware RL to the LLM to preserve the experiences. On the other hand,
recent work like Algorithm Distillation~\citep{MichaelLaskin2022_AlgorithmDistillation} presents an in-context
reinforcement learning by embedding the RL training trajectories into the input prompt of a pretrained decision transformer.
This method manages to make use of past interaction experiences without model fine-tuning.
However, existing LLMs suffer from a serious limitation of the input length
to embed the whole experience.
Hence, to better store a plethora of interaction histories and aid LLMs in 
learning during the interaction process, we introduce \textbf{RLEM}, \textit{i.e.}, \textbf{R}einforcement 
\textbf{L}earning with \textbf{E}xperience \textbf{M}emory, which accomplishes agent learning by
updating the experience memory through the RL process, rather than modifying the model parameters.

\begin{figure}[t]
    \centering
    \includegraphics[width=0.9\linewidth]{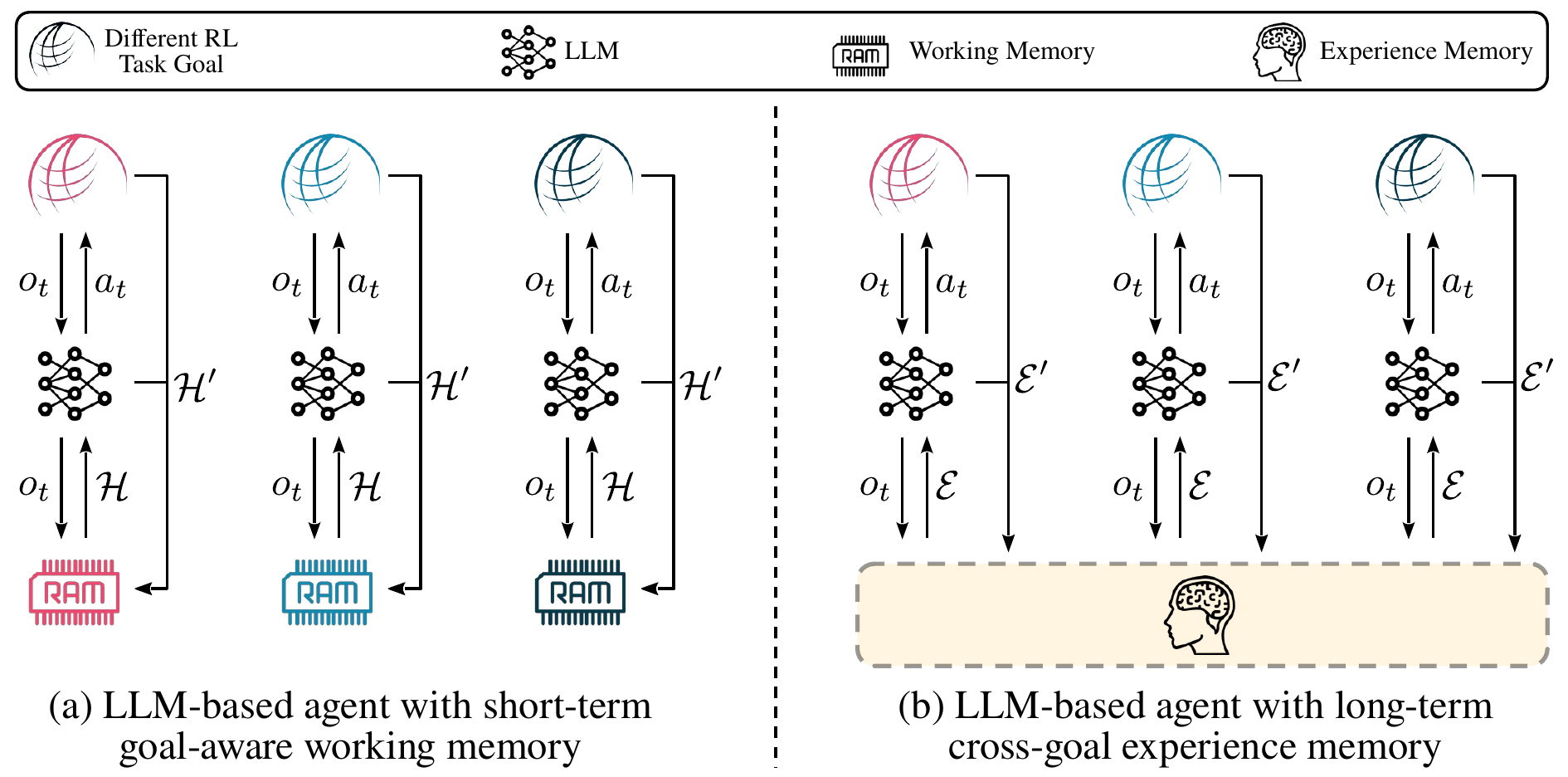}
    \caption{Comparison of the LLM-based agents with short-term working memory and
    long-term experience memory. The working memory stores only the historical
    information of the current
    episode ($\mathcal{H}$). 
    while the experience memory stores
    the interaction experiences ($\mathcal{E}$) permanently. }
    \label{fig:extnl_cmp}
\end{figure}

An external experience memory is different from the existing work like Reflexion~\citep{NoahShinn2023_Reflexion}
which combines the LLM with a short-term
working memory. As depicted in Figure~\ref{fig:extnl_cmp}~(a), a working memory is
tied to a specific task goal, and the stored histories cannot be leveraged in future episodes for different goals.
This analogy can be drawn to the Random Access Memory (RAM) of a computer, where stored information is lost 
in the event of power removal. On the other side, learning from the successful or failed experiences stored in the memory is different
from the existing work like Inner Monologue~\citep{WenlongHuang2022CoRL_InnerMonologue},
Corrective Re-Prompting~\citep{ShreyasSundaraRaman2022_CorrectiveReprompting},
and DEPS~\citep{ZihaoWang2023_DEPS} that takes advantage of immediate failure feedback only once.
Storing long-term experiences in a persistent memory gives an opportunity to discover the late failure
and learn from the experiences in the past episodes even for different task goals (see Figure~\ref{fig:extnl_cmp}~(b)).

By combining RLEM with LLM, we propose \Llmep{}, an evolvable LLM-based agent framework for decision-making tasks.
\Llmep{} can utilize the experiences
stored in the memory selectively in accordance with the current interaction state to optimize the decision.
Meanwhile, the experience memory can be updated through an RL process constantly. Such a joint system is regarded as a semi-parametric RL agent, 
which can evolve its ability through its interaction experiences analogically to a full-parametric system,
however, without fine-tuning the LLM parameters.
We evaluate \Llmep{} on two recent RL task sets with the promising performance of LLM-based agents, WebShop~\citep{ShunyuYao2022_WebShop}
and WikiHow~\citep{DanyangZhang2023_MobileEnv}. The agent is trained on a few tasks and tested on some
other tasks to check whether the experiences from different tasks can help the agent in the decision 
of the unseen episodes. 
\Llmep{} demonstrates a significant performance boost compared to both previous SOTA and our fixed-exemplar
LLM baselines.
Specifically, it achieves an average improvement of 2 points 
and 4 points on the Webshop and WikiHow tasks, respectively, compared to the SOTA models.


Our contributions are summarized as follows: 1) A new agent framework is proposed as \Llmep{}
for LLM to learn from the experiences of past episodes. The experiences are stored in an external
persistent memory instead of fine-tuning the LLM parameters or forming an extremely long prompt.
2) We introduce RLEM, which updates experience memory through analogical RL training so that
\Llmep{} is capable of self-evolving. 3) \Llmep{} manages to bypass the baseline and the prior advanced performances
and set up a new state of the art on two recent benchmarks, WebShop (+2 points on SOTA) and WikiHow (+4 points on SOTA).

\section{Related work}
\label{sec:rlt_w}

\paragraph{LLM with external information} External information is usually
adopted to augment the LLM with the environment-grounded information, or to
reduce the hallucination, or to unleash the ability to process longer context.
Connecting with an external knowledge base is a common choice for
question-answering and conversational tasks
\citep{BaolinPeng2023_LLMAugmenter,TimoSchick2023_Toolformer,HarshTrivedi2022_IRCoT,XiaomanPan2022_KiC}.
However, an external knowledge base is usually not directly corresponding to an
RL environment and cannot provide environment-grounded assistance to an
automatic agent.  Meanwhile, the update to a mature knowledge base may not be
in time for the instant interaction of the agent with the environment. In
contrast, \citet{DaleSchuurmans2023_TuringMachine} simulates a universal Turing
machine with a RAM-augmented LLM and demonstrates the capability of a
quickly-updatable working memory.  \citet{XinnianLiang2023_SCM} and
\citet{WanjunZhong2023_MemoryBank} adopt memory to store the conversational
history and handle extremely long contexts. Relational database is leveraged to
track states in a dynamic process by ChatDB~\citep{ChenxuHu2023_ChatDB}.
Reflexion~\citep{NoahShinn2023_Reflexion} exploits a working memory to store
experiences for a dedicated task to improve the performance of the agent
through several trials. However, as illustrated in Figure~\ref{fig:extnl_cmp},
the histories stored in working memory cannot benefit the episode for different
task goals. Instead, a long-term cross-goal experience memory should be
considered. MemPrompt~\citep{MadaanAman_MemPrompt} and
Ret-LLM~\citep{AliModarressi2023_RetLLM} adopt a persistent memory to store
human feedback and remind the chatbot of the conversational knowledge and
improve it continuously. Voyager~\citep{GuanzhiWang2023_Voyager} designs a
skill library to store the past learned skills as JavaScript functions. A
simple text experience pool is adopted by GITM~\citep{XizhouZhu2023_GITM} to
store the successful trajectories for future referencing. Somewhat similar to
GITM, \Llmep{} adopts a persistent environment-grounded experience memory to
store the experiences and assist in future decision-making even for different
task goal. However, instead of plain text records of successful trajectories,
\Llmep{} uses a structured memory and designs a mechanism to task advantage of
both successful and failed experiences. The experiences come from the
interaction of the agent with the environment, and no human intervention is
needed.

\paragraph{LLM learning from failure}
Learning from failure is one of the characteristic capabilities of human and turns to be an important topic for
general artificial intelligence. Some work has explored the ability of the LLM to learn from its failure
\citep{WenlongHuang2022CoRL_InnerMonologue,ShreyasSundaraRaman2022_CorrectiveReprompting,ZihaoWang2023_DEPS}.
Nonetheless, most of such work takes advantage of immediate feedback from the environment and the correction is used
only once. In practice, several late failures may be due to some early mistaken actions in an episode. 
Reflexion~\citep{NoahShinn2023_Reflexion} designs a heuristic function to detect late failure from the interaction history and 
stores the LLM-generated reflection in a working memory for use in the next trial. However, these reflections cannot be
applied to different task goals. \citet{MadaanAman_MemPrompt} stores the failure corrections for a long term, but relies on
human feedback. In contrast, \Llmep{} adopts RL to learn from both late and immediate failure from the environment
rewards without need for human feedback. Also, \Llmep{}
enables the experiences to be reused in the future episode even for a different task goal with a long-term 
experience memory.

\paragraph{LLM for decision-making} 
The powerful capability of LLM is exploited by recent 
work~\citep{WenlongHuang2022ICML_ZeroShotPlanner,ShreyasSundaraRaman2022_CorrectiveReprompting,OierMees2022_HULCpp,BoyuanChen2022_NLMap,BrianIchter2022CoRL_SayCan,WenlongHuang2022CoRL_InnerMonologue,JachyLiang2022_CaP}
to generate better control plans for various robots and agents. \citet{GeunwooKim2023_RCI} and \citet{DanyangZhang2023_MobileEnv} 
design LLM-based
agents for user interface (UI) interaction. ReAct~\citep{ShunyuYao2022_ReAct} combines the action decision with 
natural language reasoning and achieves a promising performance. To our best knowledge, This work is the first one
that combines the LLM-based agent with
RL algorithm to learn from the interaction experiences and achieve self-evolving.

The proposed \Llmep{} equips the LLM with an external experience memory to help it to learn from both 
successful and failed experiences. This is also the first work to combine the LLM-based agent with RL algorithm to
improve the capability of the agent.

\section{Method}
\label{sec:mthd}

\subsection{RLEM pipeline}

\begin{figure}[t]
	\centering
	\includegraphics[width=0.9\linewidth]{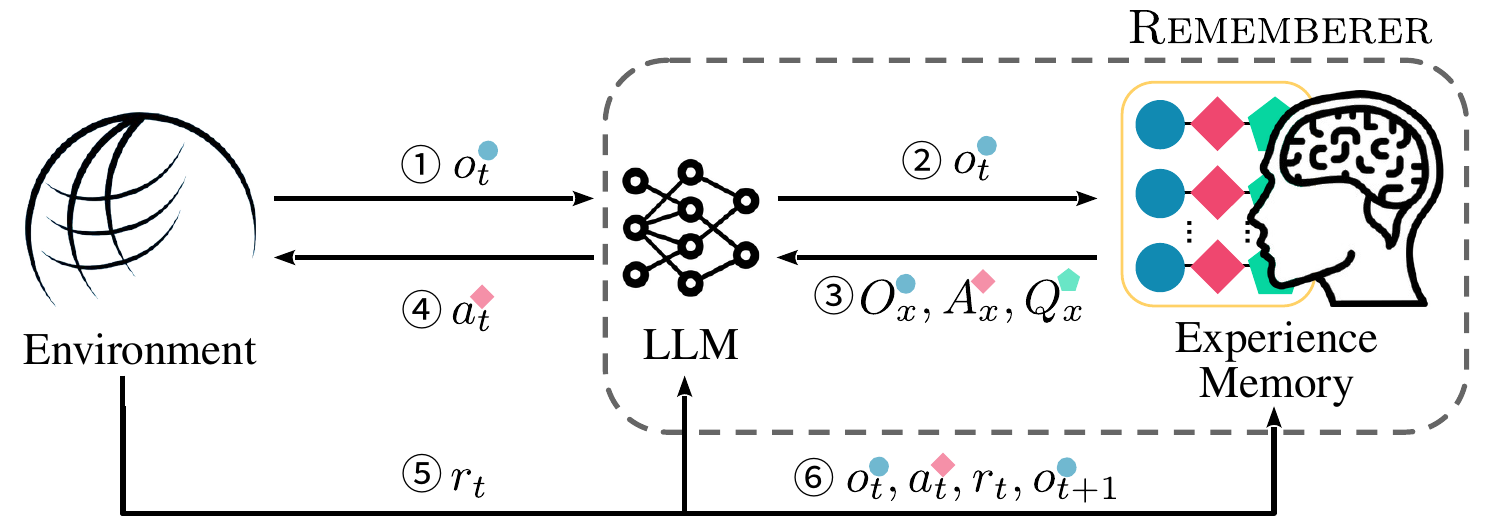}
	\caption{Pipeline of RLEM and architecture of \Llmep{}}
	\label{fig:ovrll_arch}
\end{figure}

RLEM (Reinforcement Learning with Experience Memory) is proposed for an
LLM-based agent to learn from its interaction experiences by updating an
external persistent memory.  The pipeline of RLEM and the architecture of
\Llmep{} agent are depicted in Figure~\ref{fig:ovrll_arch}.  \Llmep{} agent
consists of two components: an LLM making decisions and an experience memory
storing the interaction experiences.  At the decision step, the LLM first takes
an observation $o_t$ from the environment. The observation $o_t$ is then
adopted to retrieve several related experiences from the connected experience
memory according to some similarity functions. The experiences are represented
as a group of observations $O_x$, actions $A_x$, and the corresponding $Q$
value estimations $Q_x$. Here $x$ denotes the index set of retrieved
experiences and depends on the specific runtime observation $o_t$.
Subsequently, LLM will decide the action $a_t$ in accordance with $o_t$, the
feedback from the last interaction (\textit{e.g.}, the reward $r_{t-1}$), as
well as the retrieved experiences $(O_x, A_x, Q_x)$.  $a_t$ will be executed in
the environment and the resulted reward $r_t$ will be returned to the LLM as
the feedback. And the transition tuple, $(o_t, a_t, r_t, o_{t+1})$, comprising
the last observation, the taken action, the corresponding reward, and the new
observation will be used to update the experience memory.  The following
subsections will detail the structure and updating policy of \Llmep{}
experience memory and the usage of the retrieved experiences.

\subsection{Experience memory of \Llmep{}}
\label{sub:exp_p}

\begin{wrapfigure}{R}{0.5\linewidth}
	\centering
	\includegraphics[width=0.87\linewidth]{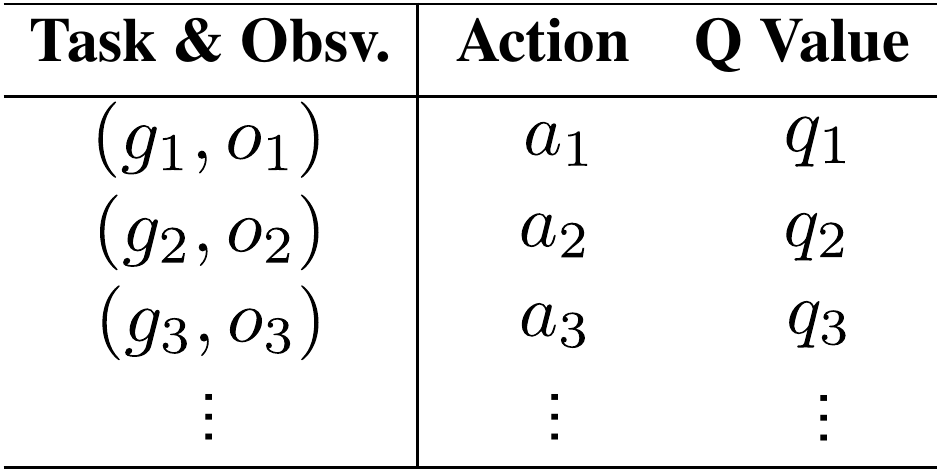}
	\caption{An example of the records stored in the proposed experience memory.}
	\label{fig:expr_p_rcd}
\end{wrapfigure}

The experience memory is one of the pivotal components of the proposed \Llmep{}
framework. It is adopted to store the interaction experiences, and the LLM is
expected to benefit from the stored experiences in future decision-making. The
memory can be regarded as a group of external parameters of the LLM-based
agent.  Such an agent is a semi-parametric system that can evolve through RL
process. During the interaction, new experiences are added to the experience
memory so that the overall system can attain a more capable interaction ability
compared to the agents with just a fixed LLM and fixed exemplars. This
procedure can be considered analogous to the training stage of conventional
parametric agents.

To be specific, the experience memory is designed as a table storing the task
information, observation, action, and the corresponding $Q$ value estimation.
The $Q$ value is the expectation of the accumulated future reward and gives an
assessment of the value of the action candidate. 
Figure~\ref{fig:expr_p_rcd} depicts a demonstration of the proposed experience
memory. There are two stages to build a practical \Llmep{} agent with
experience memory: \emph{initialization} and \emph{training}.  The experience
memory is supposed to be first initialized with some initial records before the
training stage.  The initial records are necessary to inform the LLM of the
format of the input and the output. Then, during the analogical training stage,
the agent interacts with the environment to collect new experiences, and
conducts off-policy
learning~\citep{SuttonRichardS1999Robotica_OffPolicyLearning}.  Particularly,
given the task information $g$ and the new transition $(o_t, a_t, r_t,
o_{t+1})$, as a quadruple of the last observation, action, reward, and the new
observation, a new estimation is calculated first according to the estimated
Bellman optimality equation~\citep{BellmenRichard1952NAS_BellmanEquation} as
\begin{equation}
	Q^\prime(g, o_t, a_t) = r_t + \gamma\max_a Q(g, o_{t+1}, a).
	\label{eqn:n_estmt}
\end{equation}
Here $\max$ can be calculated from the actions already recorded for $(g,
o_{t+1})$ by considering the $Q$ value of unrecorded actions 0, if the action
space cannot be traversed, \textit{e.g.}, action space involving free-form
language.  Then a new record is inserted directly if there does not exist a
record associated to $(g, o_t, a_t)$ in the memory:
\begin{equation}
	Q(g, o_t, a_t) = Q^\prime(g, o_t, a_t).
	\label{eqn:blm_eqn}
\end{equation}
If $(g, o_t, a_t)$ has been already inserted into the record, the recorded $Q$
value estimation will be updated by
Q-Learning~\citep{Watkins1992MachineLearing_Q_Learing}:
\begin{equation}
	Q(g, o_t, a_t) \leftarrow (1-\alpha) Q(g, o_t, a_t) + \alpha Q^\prime(g, o_t, a_t).
	\label{eqn:q_l}
\end{equation}
Here the learning rate, $\alpha$, is $1/N$ where $N$ denotes the times this
value is updated.  As Equation~\ref{eqn:n_estmt} may lead to an inaccurate
estimation owing to insufficient sampling out of few training steps of
\Llmep{}, $n$-step bootstrapping~\citep{VolodymyrMnih2016ICML_A3C} is adopted
to ameliorate this problem, which estimates $Q^\prime$ by
\begin{equation}
	Q^\prime(g, o_t, a_t) = \sum_{i=0}^{n-1} \gamma^ir_{t+i} + \gamma^n\max_a Q(g, o_{t+n}, a),
	\label{eqn:n_st_fltn}
\end{equation}
where $n$ is the steps to expand.  The ablation study in
Subsection~\ref{sub:ablt_std} proves this perspective.

\subsection{Usage of the experiences}
\label{sub:exp_usg}

\begin{wrapfigure}{R}{0.5\linewidth}
	\centering
	\includegraphics[width=0.95\linewidth]{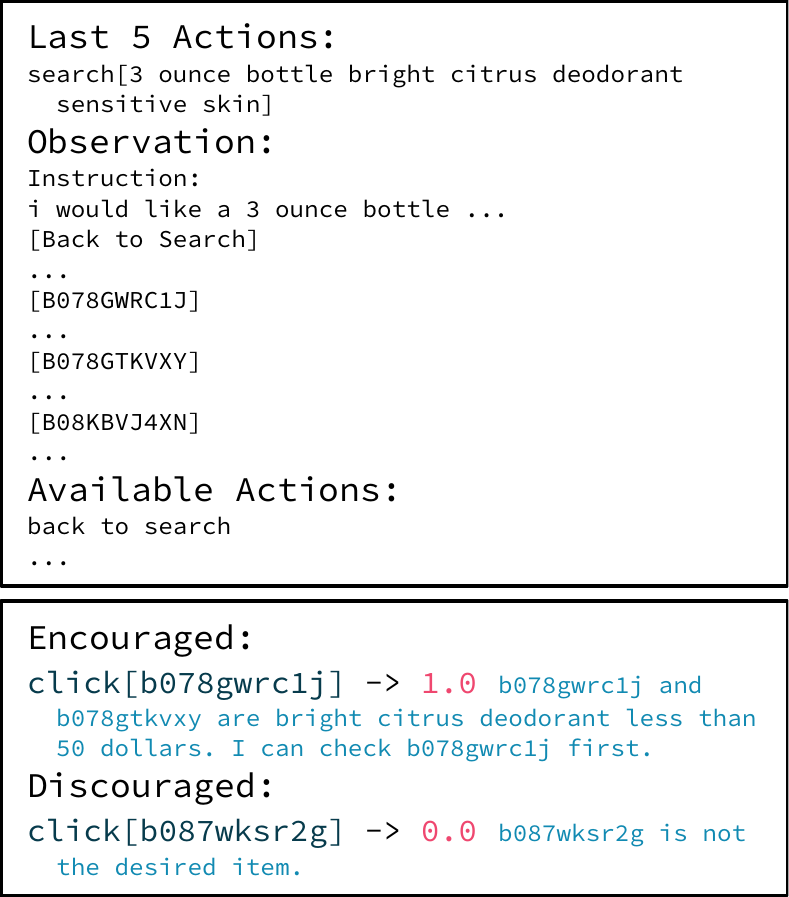}
	\caption{An exemplar for WebShop task set~\citep{ShunyuYao2022_WebShop}.
    The input part is depicted in the upper box and the output part is depicted in the 
    lower box. Action candidates are advised along with 
    their $Q$ value estimations and some optional extra information.}
\label{fig:s_crtcz}
\end{wrapfigure}

In order to assist the LLM in making decisions, the stored experiences are
adopted as dynamic exemplars for few-shot in-context learning. Given the
task goal $g$ and the current observation $o_t$, a similarity function $f$
is used to calculate the similarity of $(g, o_t)$ with $(g_i, o_i)$ from the
memory.
\begin{equation}
	S_i = f((g, o_t), (g_i, o_i)).
	\label{eqn:ccl_smlt}
\end{equation}
Commonly, a similarity function $f$ can be divided into two components, task
similarity $f_g$ and observation similarity $f_o$:
\begin{equation}
    S_i = \lambda f_g(g, g_i) + (1-\lambda) f_o(o_t, o_i).
    \label{eqn:smlt_dvd}
\end{equation}
The $m$ records with the highest similarities are retrieved to form the exemplars 
in the prompt. The particular similarity function designed 
for each task set is detailed in Subsection~\ref{sub:exp_stp_imp_dtl}.

The exemplar is supposed to demonstrate the format of the input and the output
to the LLM. The input part usually comprises the task information and the
observation, along with some interaction feedback or auxiliary information. The
particular input format depends on the task domain and will be detailed in
Subsection~\ref{sub:exp_stp_imp_dtl}.  The output part indicates the action
decision. Specifically, we propose to present the action decisions in a form of
``action advice'' comprising both encouraged and discouraged actions rather
than simply present an action to execute.  This is motivated by the perspective
that ``{\em reasoning is remembering}'' to exploit both successful and failed
experiences.  To form the output part in the exemplar, the actions with the
highest $Q$ value estimations from the retrieved record are given as the
encouraged actions, while the actions with poor $Q$ value estimations
(\textit{e.g.}, zero or negative estimations) are given as the discouraged
actions. It is believed that the advice with high value expectations can lead
the LLM to follow the past success, while the advice with poor expectations
will teach the LLM to avoid a similar failure.  A clear depiction of the
exemplar format can be found in Figure~\ref{fig:s_crtcz}.  Prompted by such
exemplars, the LLM will also predict both encouraged and discouraged actions
and speculate their $Q$ values given a new input. The predicted $Q$ values are
used to select the optimal action, to be specific, the encouraged action with
the highest $Q$ value speculation will be executed in the environment.

It is worth noting that \Llmep{} agent necessitates only a limited number of
training steps to achieve a promising performance, which leads to a
non-exhaustive action record within its memory.  Consequently, instances may
arise where there is only one action associated with a given context $(g,
o_t)$, or the highest $Q$ value remains deficient, or no sufficiently
unfavorable action exists to discourage. In such cases, randomly sampled action
advice is favored over encouraging an action with low expectations or
discouraging an action with moderate expectations. Our ablation study in
Subsection~\ref{sub:ablt_std} sheds light on various strategies for generating
advice in such scenarios.

\section{Experiments \& results}
\label{sec:exp_rslt}

\subsection{Experiment setup \& implementation details}
\label{sub:exp_stp_imp_dtl}

To assess the effectiveness of \Llmep{}, we evaluate it on two recent task sets with the promising performance of
LLM-based agents: WebShop and WikiHow.
All the experiments are conducted based on the OpenAI API of 
GPT-3.5~\citep{TomBBrown2020NeurIPS_GPT3} text-davinci-003\footnote{\url{https://openai.com/api/}}.

\paragraph[WebShop]{WebShop~\citep{ShunyuYao2022_WebShop}} WebShop is a task
set simulating a web store site. The agent is instructed to browse the site and
shop for the target goods. The information of over 1M products is crawled from
the Amazon store\footnote{\url{https://www.amazon.com/}}. About 12K product
requests are re-written by crowd laborers to generate more diverse
instructions. A score between 0 and 1 will be rated after shopping by assessing
the correspondence between the product and the instruction. We followed
\citet{NoahShinn2023_Reflexion} and conducted our experiments on the first 100
tasks from the same shuffled task list released along with the task set. At
each interaction step, the LLM takes the web page representation and a list of
available actions as input. The task instruction is omitted, for there is
always an instruction present on the top of the web page. As there are no
intermediate rewards during the episode, only the last 5 performed actions
serve as procedure feedback.  Inspired by the chain-of-thought
technique~\citep{JasonWei2022_CoT} and the ReAct
mechanism~\citep{ShunyuYao2022_ReAct}, the LLM is prompted to predict a reason
for its decision as the extra information depicted in Figure~\ref{fig:s_crtcz}.
The representation of the web pages is simplified in the same way as ReAct. The
task similarity $f_g$ is calculated using the all-MiniLM-L12-v2 model from
Sentence-Transformers~\citep{NilsReimers2019EMNLP_IJCNLP_SentenceTransformers}.
As it is noticed that the web pages in WebShop are instantiated from some
templates, we categorize the web pages into four patterns and design a
similarity lookup table to compute the observation similarity $f_o$ according
to the web page patterns.  The details about the similarity table should be
referred to in the supplementary.  It is observed that most of the tasks end in
5 steps, thus we directly conduct a full-trajectory expanding while performing
multi-step bootstrapping:
\begin{equation}
    Q^\prime(o_t, a_t) = \sum_{\tau=t}^{T} \gamma^{\tau-t}r_\tau.
    \label{eqn:f_trjctr_flt}
\end{equation}

\paragraph[WikiHow]{WikiHow~\citep{DanyangZhang2023_MobileEnv}} WikiHow is a
task set based on the collaborative wiki app
WikiHow\footnote{\url{https://www.wikihow.com/Main-Page}} running on the
interaction platform Mobile-Env~\citep{DanyangZhang2023_MobileEnv}. The task
set contains amounts of navigation tasks. The target of the agent is to follow
the instructions and navigate to the required page.  Intermediate rewards and
instructions may be triggered during the episode.  We followed
\citet{DanyangZhang2023_MobileEnv} and evaluated the proposed \Llmep{} on the
``canonical subset'' comprising 70 tasks. Specifically, the LLM is input with
the task description, the screen representation, and the step instruction. The
screen is represented in an HTML element sequence following
\citet{DanyangZhang2023_MobileEnv}.  Additionally, the last 5 performed actions
along with the last reward are given to the LLM as the procedure feedback. As
for the output, the LLM is prompted to print the HTML representation of the
operated element as the extra information.  This is expected to force the LLM
to discover the relation between the element id and the certain element.  The
task similarity $f_g$ designed for WikiHow is computed from the step
instructions. It is noticed that the instructions follow some patterns, thus,
we inspect the instructions and categorize them into six types. Then a
similarity lookup table is designed according to the instruction types.  The
details should be referred to in the supplementary. The observation similarity
$f_o$ is computed based on the length of the longest common sequence of the
HTML elements in the screen representation:
\begin{equation}
    f_o(sc_1, sc_2) = \dfrac{lcs(sc_1, sc_2)}{\max\{len(sc_1), len(sc_2)\}}.
    \label{eqn:lcs_sm}
\end{equation}
The full-trajectory expanding is adopted, as most of the tasks will end in 5
steps as well.

\begin{table}[t]
    \begin{minipage}{0.49\linewidth}
        \centering
        \caption{Results on WebShop. The result of the prior state of the art, 
        ReAct~\citep{ShunyuYao2022_ReAct}, is attained 
        with the public implementation released by the original authors.
        The RL, IL, and IL+RL results are retrieved directly
        from \citet{ShunyuYao2022_WebShop}.}
        \label{tab:webshop1}
        \begin{tabular}{ccc}
            \toprule[1.5pt]
            Method & Avg Score & Success Rate \\
            \midrule
            LLM only &  0.55 &  0.29 \\
            ReAct &     0.66 &  0.36 \\
            \textbf{\llmep{}} &  \textbf{0.68} &  \textbf{0.39} \\
            \midrule
            RL &    0.55 &  0.18 \\
            IL &    0.60 &  0.29 \\
            IL+RL & 0.62 &  0.29 \\
            \bottomrule[1.5pt]
        \end{tabular}
    \end{minipage}
    ~
    \begin{minipage}{0.49\linewidth}
        \centering
        \caption{Results on WikiHow. ``Mobile-Env'' indicates the prior result from 
        \citet{DanyangZhang2023_MobileEnv}.
        ``\llmep{} (A)'' denotes the results
        by directly running the evaluation of \Llmep{} with a human-annotated
        experience memory.
        }
        \label{tab:wikihow1}
        \begin{tabular}{ccc}
            \toprule[1.5pt]
            Method & Avg Reward & Success Rate \\
            \midrule
            LLM only &  2.58 &  0.90 \\
            Mobile-Env & 2.50 &    0.89 \\
            \textbf{\llmep{}} &  \textbf{2.63} &  \textbf{0.93} \\
            \midrule
            \llmep{} (A) & 2.56 &   0.91 \\
            \bottomrule[1.5pt]
        \end{tabular}
    \end{minipage}
\end{table}

\begin{table}[t]
    \centering
    \caption{Results on WebShop with different exemplar combinations (initial experiences
    for \Llmep{}) and different training sets (for \Llmep{}). 
    $E_i$ denotes the different exemplar combinations,
    while $S_i$ denotes the different training sets.
    The first line of each method shows the mean scores, and the second line shows the success rates.
    }
    \label{tab:webshop2}
    \begin{tabular}{ccccccc}
        \toprule[1.5pt]
         &  &   \multicolumn{2}{c}{Different (Initial) Exemplars} & Different Training Sets  & & \\
         \cmidrule(lr){3-4}\cmidrule(lr){5-5}
         & $E_0+S_0$ &   $E_1+S_0$  & $E_2+S_0$ &   $E_0+S_1$ & Avg & Std \\
         \midrule
         \multirow{2}{*}{ReAct} &    \textbf{0.72} & 0.65 &  0.60 &  - & 0.66 & 0.06 \\
         & \textbf{0.42} &  0.35 &  0.30 &   - & 0.36 & 0.06 \\
         \midrule
         \multirow{2}{*}{LLM only} &    0.52 &  0.54 & 0.59 & - & 0.55 & 0.04 \\
         &  0.26 &  0.28 & 0.32 & -   & 0.29 &    0.03 \\
         \midrule
         \multirow{2}{*}{\llmep{}} &   0.66 &  \textbf{0.71} &  \textbf{0.66} &   \textbf{0.67} & \textbf{0.68} &  \textbf{0.02} \\
         &  0.37 &  \textbf{0.41} &  \textbf{0.37} &  \textbf{0.40} &  \textbf{0.39} & \textbf{0.02} \\
         \bottomrule[1.5pt]
    \end{tabular}
\end{table}

\begin{table}[t]
    \begin{minipage}{0.49\linewidth}
        \centering
        \caption{Comparison of the number of annotated trajectories and steps of \Llmep{} and the IL
        baseline. The number of steps of the training set of IL is estimated according to the average 
        human trajectory length on the test split as 11.3 in \citet{ShunyuYao2022_WebShop}.}
        \label{tab:annt_trj_st_n}
        \begin{tabular}{ccc}
            \toprule[1.5pt]
            Method &    \#Trajectories & \#Steps \\
            \midrule
            IL &    1,012 & \textasciitilde 11,436 \\
            \Llmep{} & 1 &  4 \\
            \bottomrule[1.5pt]
        \end{tabular}
    \end{minipage}
    ~
    \begin{minipage}{0.49\linewidth}
        \centering
        \caption{Comparison of the number of the tasks in the training set and the updating steps of \Llmep{}
        with the IL and RL baselines. The number of the updating steps of IL is estimated from 10 epochs on
        1,012 trajectories with an average trajectory length of 11.3.}
        \label{tab:tr_t_st_n}
        \begin{tabular}{ccc}
            \toprule[1.5pt]
            Method &    \#Tasks &    \#Steps \\
            \midrule
            RL &    10,587 &    100,000 \\
            IL &    - & \textasciitilde 114,356 \\
            \Llmep{} & 10 & 74 \\
            \bottomrule[1.5pt]
        \end{tabular}
    \end{minipage}
\end{table}

\begin{table}[t]
    \centering
    \caption{Results on WikiHow with different exemplar combinations (initial experiences
    for \Llmep{}) and different training sets (for \Llmep{}). 
    }
    \label{tab:wikihow2}
    \begin{tabular}{ccccccc}
        \toprule[1.5pt]
         &  &   \multicolumn{2}{c}{Different (Initial) Exemplars} & Different Training Sets &   & \\
         \cmidrule(lr){3-4}\cmidrule(lr){5-5}
         & $E_0+S_0$ &  $E_1+S_0$ & $E_2+S_0$ &   $E_0+S_1$ & Avg & Std \\
         \midrule
         \multirow{2}{*}{LLM only} &    2.56 &  2.60 & \textbf{2.59} & - & 2.58 &  \textbf{0.02} \\
         &  0.90 &  0.90 & 0.89 & - & 0.90 &  \textbf{0.01} \\
         \midrule
         \multirow{2}{*}{\llmep{}} &    \textbf{2.63} &  \textbf{2.63} &   \textbf{2.59} &  \textbf{2.66}   & \textbf{2.63} & 0.03 \\
         &  \textbf{0.93} &  \textbf{0.91} &  \textbf{0.90} &  \textbf{0.97} &  \textbf{0.93} & 0.03 \\
         \bottomrule[1.5pt]
    \end{tabular}
\end{table}

\subsection{Results on WebShop}

\Llmep{} is applied to WebShop with 2-shot in-context learning. The experience
memory is initialized with four annotated experiences of the decision step from one trajectory. 
The agent is trained for 3 epochs on a training
set containing 10 different tasks outside the test sets used by \citet{ShunyuYao2022_ReAct} and \citet{NoahShinn2023_Reflexion}.
To control the total expense and achieve bootstrapping, the succeeded tasks in the first
epoch are excluded from training in the following two epochs. 
The trajectories exceeding 15 steps are considered to be failed, as most of the tasks can end in
5 steps. The main results are shown in Table~\ref{tab:webshop1}. We used the public ReAct~\citep{ShunyuYao2022_ReAct}
implementation released by the authors and run with text-davinci-003 instead of text-davince-002 in \citet{ShunyuYao2022_ReAct}.
The run of ReAct shares the same trajectory as the exemplar with \Llmep{}. 
The ``LLM only'' baseline indicates a single LLM with 2 fixed exemplars
sampled from the initial
experiences of \Llmep{}. The average performance of \Llmep{} exceeds the baseline
by a large extent and surpasses the prior state of the art, ReAct, as well. This proves the effectiveness of augmenting
the LLM with an external 
evolvable experience memory.
The proposed \Llmep{} also
outperforms the RL, IL (imitation learning), and IL+RL baselines on both metrics. 

In order to verify the robustness of \Llmep{}, experiments with different
initial experience combinations or a different training set are conducted.
The results are depicted in Table~\ref{tab:webshop2}. 
The initial experience combination $E_0$
denotes the certain trajectory adopted by the original implementation of ReAct while $E_1$ and $E_2$ are randomly
sampled from $S_0$. It is observed that the proposed \Llmep{} can achieve better and more stable results with
different initialization and training sets compared to ReAct. Thus, \Llmep{} can mitigate the workload to some extent
to search for an optimal exemplar combination.

We compare the training efficiency of \Llmep{} with the conventional IL and RL methods in Table~\ref{tab:annt_trj_st_n}
and Table~\ref{tab:tr_t_st_n}. In contrast to the IL, \Llmep{} requires quite few annotated samples to 
initialize the experience memory, while IL is in need of much more human annotations. \Llmep{} agent can be trained
on only 10 tasks for 74 steps, while the RL and IL are expected to be trained for about 100 thousand steps to achieve
an acceptable performance. Consequently, the proposed \Llmep{} offers a much more efficient way to build a practical
agent agilely.

\subsection{Results on WikiHow}

\Llmep{} is applied to WikiHow with 2-shot in-context learning. The experience memory is initialized with 
two annotated experiences of the decision step. The agent is trained for 3 epochs on a training
set containing 10 different tasks selected from WikiHow excluding the  
test tasks. 
Similar to the experiments on WebShop, the succeeded tasks in the first epoch are excluded from training in 
the following two epochs. 
As observed that most of the tasks require an interaction
of less than 5 steps, the trajectory exceeding 15 steps will be regarded as failed. The main results are depicted
in Table~\ref{tab:wikihow1}. The 
exemplars of the ``LLM only'' baseline are the initial experiences of \Llmep{}. 
The proposed \Llmep{} surpasses the baseline as well
as the original result in \citet{DanyangZhang2023_MobileEnv}. In addition, 10 tasks
are annotated to form an annotated experience memory. 
\Llmep{} agent with this annotated experience memory is evaluated without further training, and the result
is denoted as ``\llmep{} (A)'' in the table.
This result demonstrates that
\Llmep{} is capable of exploiting expert experiences, which can be regarded as analogous to conventional
imitation learning. Nevertheless, the annotated experiences may not offset the exact shortage of the
particular LLM. In contrast, the RL training will have an opportunity to collect more specific experiences
and gain a more promising performance.

The experiments with different initial experience combinations or a different training set are conducted
on WikiHow as well, and the results are shown in Table~\ref{tab:wikihow2}.
The proposed \Llmep{} achieves a consistent improvement compared to
the baseline with fixed exemplars, which proves
the effectiveness and robustness of \Llmep{}.

\subsection{Ablation study}
\label{sub:ablt_std}

Several ablation studies are conducted to verify the design of \Llmep{} framework.

\begin{table}[t]
    \centering
	\caption{Comparison of the average reward estimation of the full model and
	the ablation model without bootstrapping policy. The error is the absolute
difference between the average reward estimation from the experience memory and
the real training reward.}
    \label{tab:rwd_est_err}
    \begin{tabular}{ccccc}
        \toprule[1.5pt]
        Task Set & Setting &    Avg Reward Estimation & Avg Training Reward &   Abs Error \\
        \midrule
        \multirow{2}{*}{WebShop} &  Full Model &    0.86 &  0.84 &  \textbf{0.02} \\
        &   w/o bootstrp. &  0.62 &  0.84 &  0.22 \\
        \midrule
        \multirow{2}{*}{WikiHow} & Full Model & 2.48 &  2.60 &  \textbf{0.12} \\
        &   w/o bootstrp. &   1.98 &  2.70 &  0.72 \\
        \bottomrule[1.5pt]
    \end{tabular}
\end{table}

\begin{table}[t]
    \centering
    \caption{Results of ablation study}
    \label{tab:ablt_std}
    \begin{tabular}{cccc}
        \toprule[1.5pt]
        Task Set & Setting &    Avg Reward/Score &  Success Rate \\
        \midrule
		\multirow{3}{*}{WebShop} &  Full Model &    \uline{0.66} &  \uline{0.37} \\
         &  w/o bootstrp. &  0.67 & 0.36 \\
         &  w/o random &    0.65 &  0.37 \\
         \midrule
		 \multirow{6}{*}{WikiHow} & Full Model & \uline{2.63} &  \uline{0.93} \\
         &  w/o bootstrp. &  2.54 &  0.89 \\
         &  w/o random &    2.64 & 0.90 \\
		 &	w/o discouraged &	2.48 &	0.81 \\
		 &	w/o task sim. $f_g$ &	2.63 &	0.94 \\
		 &	w/o obsrv. sim. $f_o$ &	2.47 &	0.87 \\
        \bottomrule[1.5pt]
    \end{tabular}
\end{table}

\paragraph{Ablation on $n$-step bootstrapping policy} Ablation studies are
conducted to verify the necessity of $n$-step bootstrapping policy to update
the $Q$ value estimations in the experience memory. As stated in
Subsection~\ref{sub:exp_p}, updating without bootstrapping may lead to
inaccurate value estimations owing to few training steps to explore and
exploit. In order to verify this perspective, an average reward estimation is
calculated by averaging the sum of the maximum $Q$ value and the history reward
stored for each observation in the experience memory:
\begin{equation}
    \hat{R} = \dfrac{1}{M}\sum_{i=1}^M (R_h(g_i, o_i) + \max_a Q(g_i, o_i, a)),
\end{equation}
where $R_h$ denotes the total reward of the steps before $(g_i, o_i)$ on the
trajectory and $M$ is the size of the memory.  The deduced average reward
estimation $\hat{R}$ is compared to the real training reward, and an absolute
error is calculated in Table~\ref{tab:rwd_est_err}.  It can be observed that
the reward estimation from the experience memory trained without bootstrapping
suffers a far greater error than that with bootstrapping.  Meanwhile, the
performance on the test set is demonstrated in Table~\ref{tab:ablt_std}.
Although there is no apparent disparity in the final performance on the WebShop
task set, a visible degradation is observed on WikiHow, which reveals the
latent risk of a non-bootstrapping update.

\paragraph{Ablation on the advice generation strategy} As stated in
Subsection~\ref{sub:exp_usg}, owing to the non-exhaustive exploration in the
brief training stage, there may be no suitable candidates for the action advice
in the exemplars. For instance, there may be no actions recorded with a poor
enough $Q$ value estimation or no actions recorded as high-reward. Under this
case, action advice can be generated with a randomly sampled action that is not
in the record, or it can be given by directly encouraging the action with the
highest $Q$ value estimation and discouraging the action with the lowest
estimation without regard to the certain value. These two strategies are
compared in Table~\ref{tab:ablt_std}.  As the results illustrate, the random
plan appears to have a minor superiority over the non-random plan. This is
attributed to that advice with improper value expectations will mislead the LLM
to take wrong judgments about the true value of the available actions.

Additional experiments are conducted to investigate the necessity of the
discourage actions in the output part of exemplars and the impact of similarity
function components. Owing to limit of budgets, these experiments are only
conducted on WikiHow task set.

\paragraph{Ablation on necessity of the discouraged actions} The proposed
output format ``action advice'' comprises both encouraged and discouraged
actions.  The discouraged actions are believed to help the LLM to avoid similar
failures.  Results in Table~\ref{tab:ablt_std} prove necessity of the
discouraged actions. Without access to the discouraged actions, the agent can
only achieve a much poorer performance than the full model. In the case shown
in the supplementary, it can be seen that there may not be proper actions to
encourage in the retrieved experience. In such cases, the discouraged actions
are especially crucial for the agent to prevent repeating similar mistakes.

\paragraph{Ablation on the similarity function} As stated in
Subsection~\ref{sub:exp_usg}, a similarity function is required to select
related experiences from the memory. In experiments, the similarity is
implemented as two components: task similarity $f_g$ and observation similarity
$f_o$. Ablation studies are conducted to draw a brief perspective on the impact
of these two components. As shown in Table~\ref{tab:ablt_std}, removal of task
similarity seems not to affect the performance remarkably, while removal of
observation similarity causes a serious degradation. This may indicate that on
these tasks, the tested LLM benefits more from experiences that have similar
observations rather than similar instruction patterns. On the other side, the
pattern-based task similarity for WikiHow introduced in
Subsection~\ref{sub:exp_stp_imp_dtl} may be too coarse to cluster the
experiences. During interaction, the agent may receive instructions of the same
pattern (\textit{e.g.}, ``access article ABC'') while facing different types of
observation (\textit{e.g.}, search result page or category page). The
appropriate actions in two situations are also different.  Removal of
observation similarity will eliminate this difference in experience selection
and results in misleading. Case study in the supplementary shows this
perspective.

\section{Conclusion}
\label{sec:ccls}

We introduce Reinforcement Learning with Experience Memory (RLEM) to aid the
LLM in learning from its interaction experiences for decision-making tasks.  A
novel LLM-based agent framework called \LLMEP{} is then designed with RLEM by
equipping the LLM with a persistent experience memory and updating the memory
with the RL algorithm.  \Llmep{} agent is capable of exploiting the interaction
experiences to improve its policy and gains a significant improvement compared
to the baseline. Our experimental results demonstrate the superiority.  Owing
to the simplicity and effectiveness of \Llmep{}, we believe that this work
provides a valuable perspective on designing evolvable LLM-based agents with
RLEM.

\section{Limitations}
\label{sec:lmtt}

The proposed \Llmep{} agent demonstrates strong superiority on the tested
benchmarks. Nevertheless, it is wondered how this framework will be applied to
the environments with more long-term episodes or with more extensive or
visual-rich observations. Besides, it is observed that the performance of
\Llmep{} will encounter quick saturation in training process. This may be due
to the limited number of active exemplars. Further efforts are expected to be
dedicated in to make the agent performance evolve continuously. Furthermore, as
an early exploration, we didn't make use of complicated RL techniques. How
recent advancement in RL domain works under RLEM is also an interesting
problem.



\section*{Acknowledgements}

This work is funded by the China NSFC Project (No.62106142 and No.62120106006)
and Shanghai Municipal Science and Technology Major Project (2021SHZDZX0102).

\bibliographystyle{plainnat}
\bibliography{neurips_2023}

\appendix

\section{Details about the observation formats}

\begin{figure}[h]
	\centering
	\includegraphics[width=\linewidth]{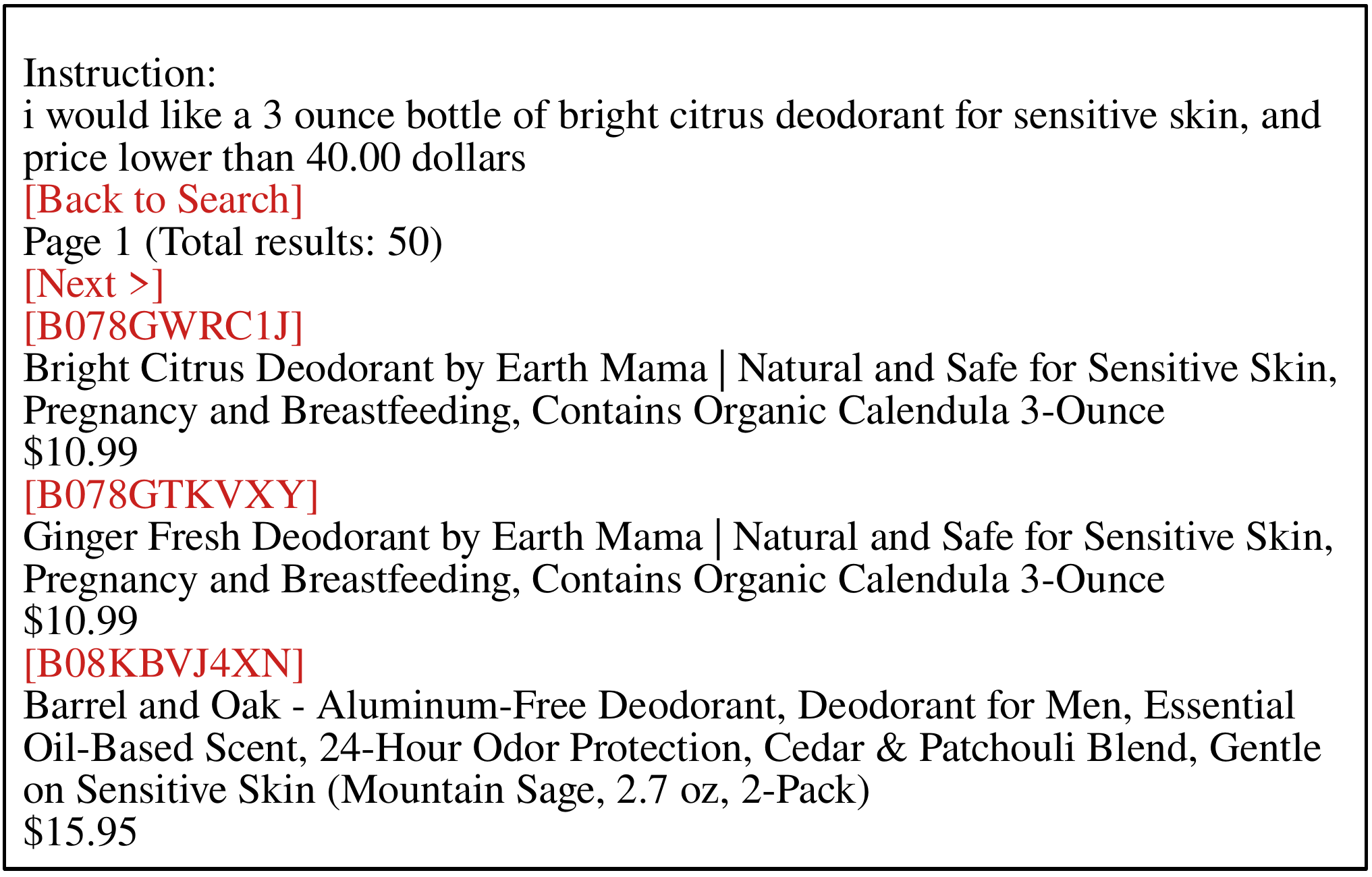}
	\caption{Example of the observation of WebShop}
	\label{fig:wsh_obsv}
\end{figure}

The observation of WebShop is simplified based on the \verb|text_rich| format
of WebShop~\citep{ShunyuYao2022_WebShop} in exactly the same way with
\citet{ShunyuYao2022_ReAct}. Specifically, the HTML markups are omitted, and
the buttons are represented in \verb|[text]| or \verb|[[text]]| instead of the
complicated \verb|[button] text [button_]| or {\tt [clicked button] text
[clicked button\_]}.  Furthermore, the number of displayed search results per
page is clipped to 3 instead of 10. An example is shown in
Figure~\ref{fig:wsh_obsv}.

The observation of WikiHow is represented in exactly the same way with
\citet{DanyangZhang2023_MobileEnv}. Specifically, the page is converted into a
sequence of HTML elements corresponding to the visible leaf nodes on the
Android~\texttrademark{} view hierarchy (VH). The node classes are converted
into HTML tags and a few VH properties are converted into similar HTML
attributes. The \verb|text| property is converted to the text content of the
common HTML element or the \verb|value| attribute of the \verb|input| element.

\section{Lookup table of the pattern-based similarity functions}

\subsection{Lookup table of the page similarity function of WebShop}

We inspected the pages from WebShop and categorized them into 4 patterns as
depicted in Table~\ref{tab:wsh_ptn}.

\begin{table}[h]
	\centering
	\caption{Patterns of WebShop pages}
	\label{tab:wsh_ptn}
	\begin{tabular}{cl}
		\toprule[1.5pt]
		Pattern &	Description \\
		\midrule
		search &	The page to search for an item \\
		itemlisting &	The page listing the search results \\
		item &	The information page of a specific item \\
		others &	The item description page, item feature page, and review page \\
		\bottomrule[1.5pt]
	\end{tabular}
\end{table}

\begin{table}[h]
	\centering
	\linespread{1.2}
	\renewcommand\arraystretch{1.2}
	\caption{Lookup table of the page similarity of WebShop}
	\label{tab:wsh_sml_tbl}
	\begin{tabular}{c|cccc}
		 &	search &	itemlisting &	item &	others \\
		\hline
		search &	1 &	0 &	0 &	0 \\
		itemlisting &	0 &	1 &	0 &	0 \\
		item &	0 &	0 &	1 &	0.3 \\
		others &	0 &	0 &	0.3 &	1 \\
	\end{tabular}
\end{table}

The similarity lookup table is defined in Table~\ref{tab:wsh_sml_tbl}.

\subsection{Lookup table of the instruction similarity function of WikiHow}

We inspected the step instructions from WikiHow and categorized them into 6
patterns as depicted in Table~\ref{tab:wkh_ptn}.

\begin{table}[h]
	\centering
	\caption{Patterns of WikiHow instructions}
	\label{tab:wkh_ptn}
	\begin{tabular}{cl}
		\toprule[1.5pt]
		Pattern Name &	Pattern Template \\
		\midrule
		search &	Search an article to learn \dots \\
		article &	Access the article \dots \\
		author &	Check the author page of \dots \\
		category &	Access the page of category \dots \\
		reference &	Check the reference list. \\
		about &	Access the about page \dots \\
		\bottomrule[1.5pt]
	\end{tabular}
\end{table}

\begin{table}[h]
	\centering
	\linespread{1.2}
	\renewcommand\arraystretch{1.2}
	\caption{Lookup table of the instruction similarity of WikiHow}
	\label{tab:wkh_sml_tbl}
	\begin{tabular}{c|cccccc}
		 &	search &	article &	author &	category &	reference &	about \\
		\hline
		search &	1 &	0.1	&	0 &	0 &	0 &	0 \\
		article &	0.1 &	1 &	0.3 &	0.3 &	0 &	0 \\
		author &	0 &	0.3 &	1 &	0.8 &	0.3 &	0.3 \\
		category &	0 &	0.3 &	0.8 &	1 &	0.3 &	0.3 \\
		reference &	0 &	0 &	0.3 &	0.3 &	1 &	0.8 \\
		about &	0 &	0 &	0.3 &	0.3 &	0.8 &	1\\
	\end{tabular}
\end{table}

The similarity lookup table is defined in Table~\ref{tab:wkh_sml_tbl}.

\section{Hyper-parameters}

The discount factor $\gamma$ to accumulate the rewards in the formula of $Q$
value is 1, which means no discounts are considered. The learning rate $\alpha$
is $1/N$ where $N$ denotes the times the value is updated. Such a learning rate
is chosen, as the tested environments are stationary and each estimation to the
value is expected to be equally weighted. The similarity weight factor
$\lambda$ is 0.5, indicating two parts of the similarity function contribute
equally.

\section{Capability evolving of \Llmep{}}

\begin{figure}[h]
	\centering
	\begin{subfigure}[b]{0.49\textwidth}
		\centering
		\includegraphics[width=\linewidth]{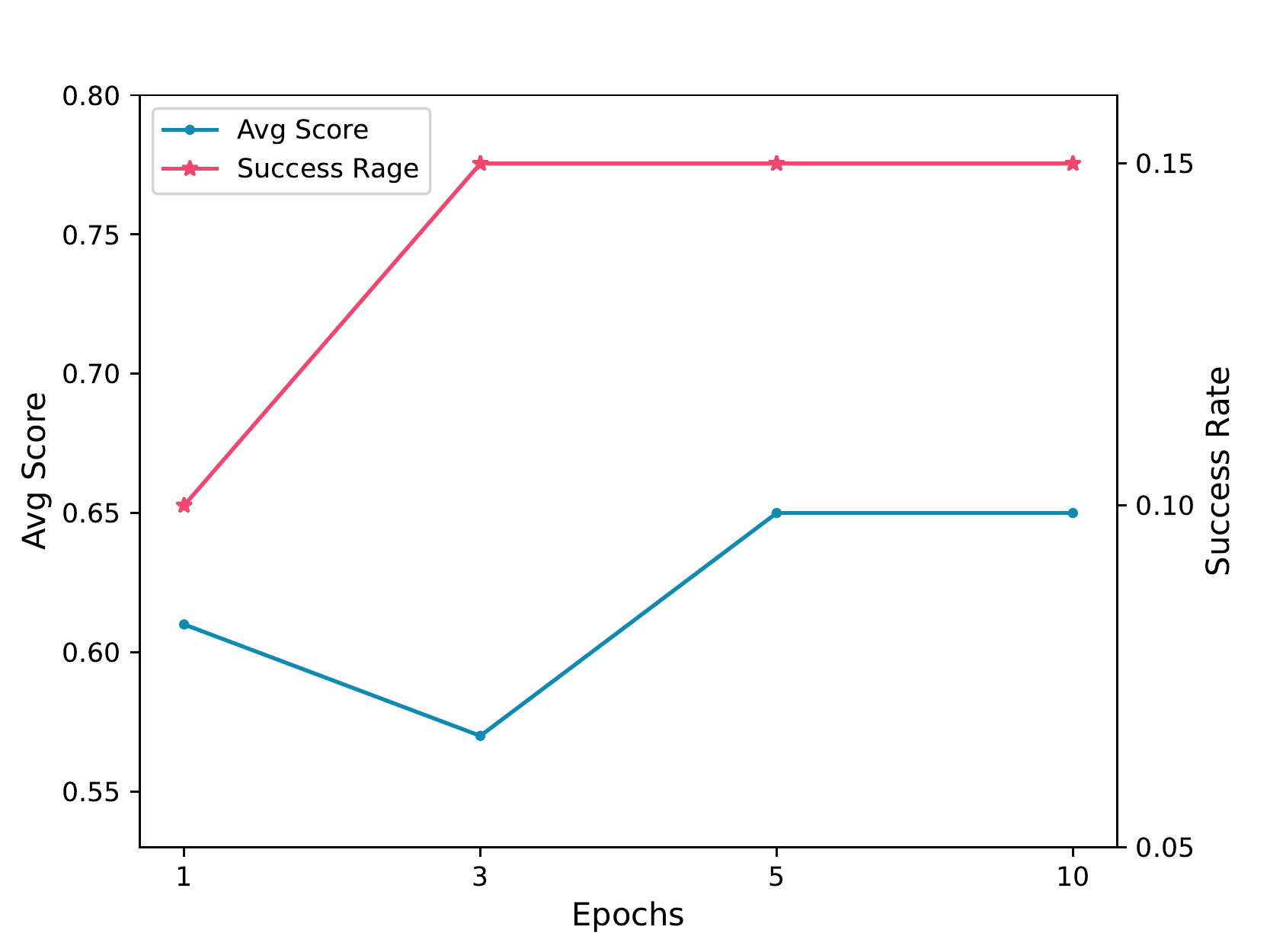}
		\caption{Capability evolving on WebShop}
	\end{subfigure}
	~
	\begin{subfigure}[b]{0.49\textwidth}
		\centering
		\includegraphics[width=\linewidth]{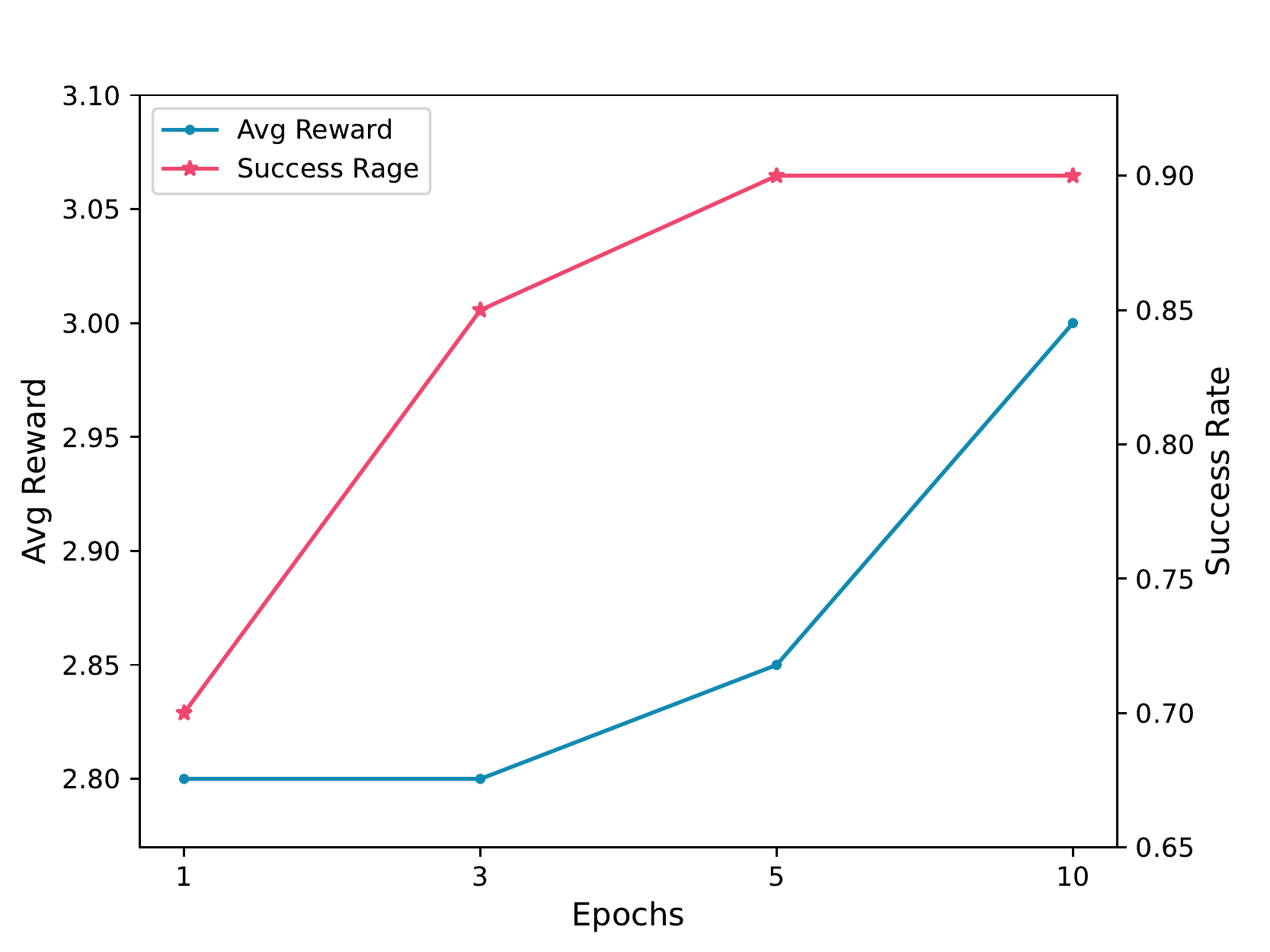}
		\caption{Capability evolving on WikiHow}
	\end{subfigure}
	\caption{Performance on a random subset at epochs 1, 3, 5, and 10}
	\label{fig:cpbl_evlv}
\end{figure}

\begin{figure}[h]
	\centering
	\begin{subfigure}[t]{0.47\textwidth}
		\begin{center}
			\includegraphics[width=\linewidth]{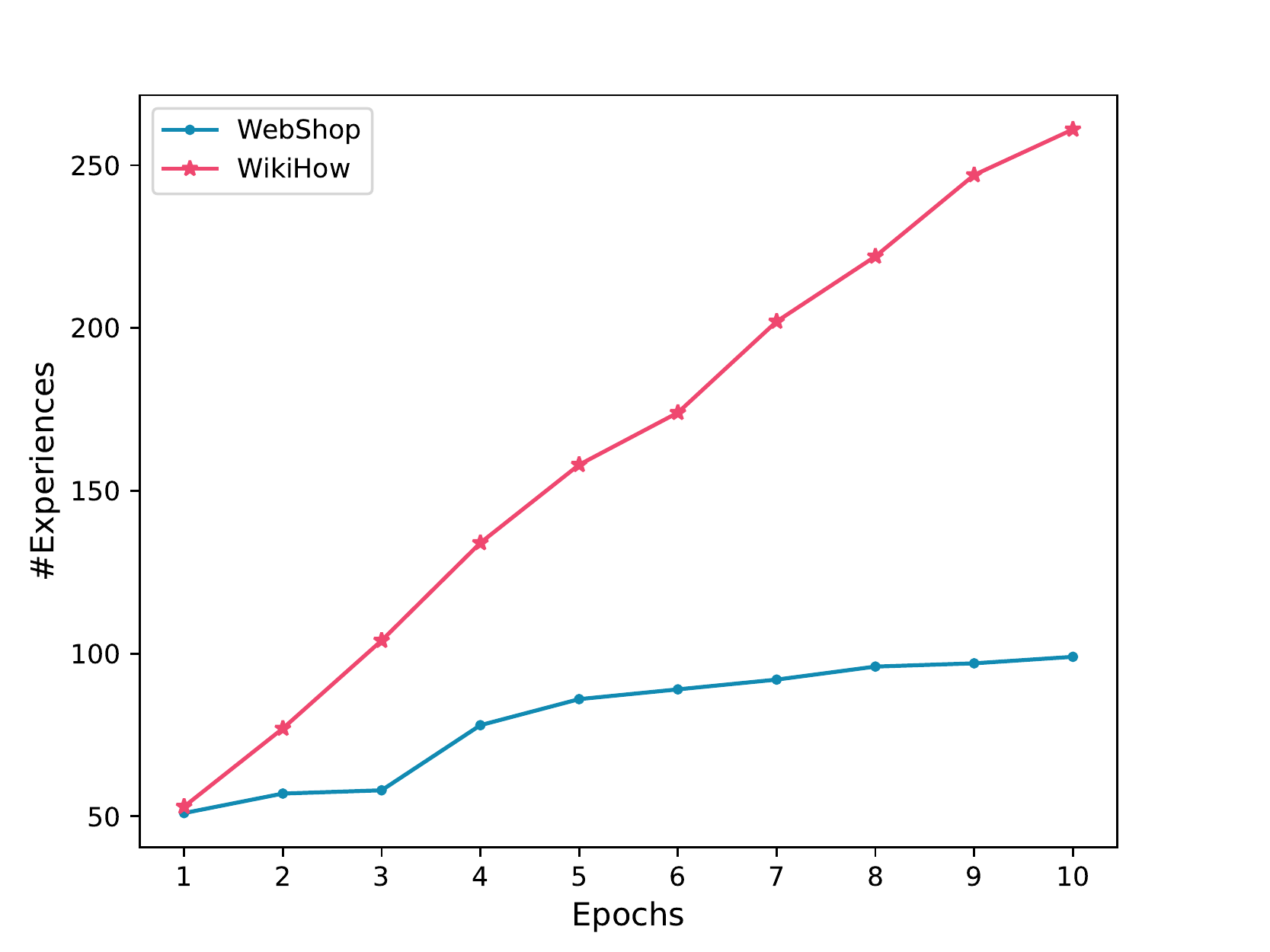}
		\end{center}
		\caption{Number of experiences in each training epoch}
	\end{subfigure}
	~
	\begin{subfigure}[t]{0.47\textwidth}
		\begin{center}
			\includegraphics[width=\linewidth]{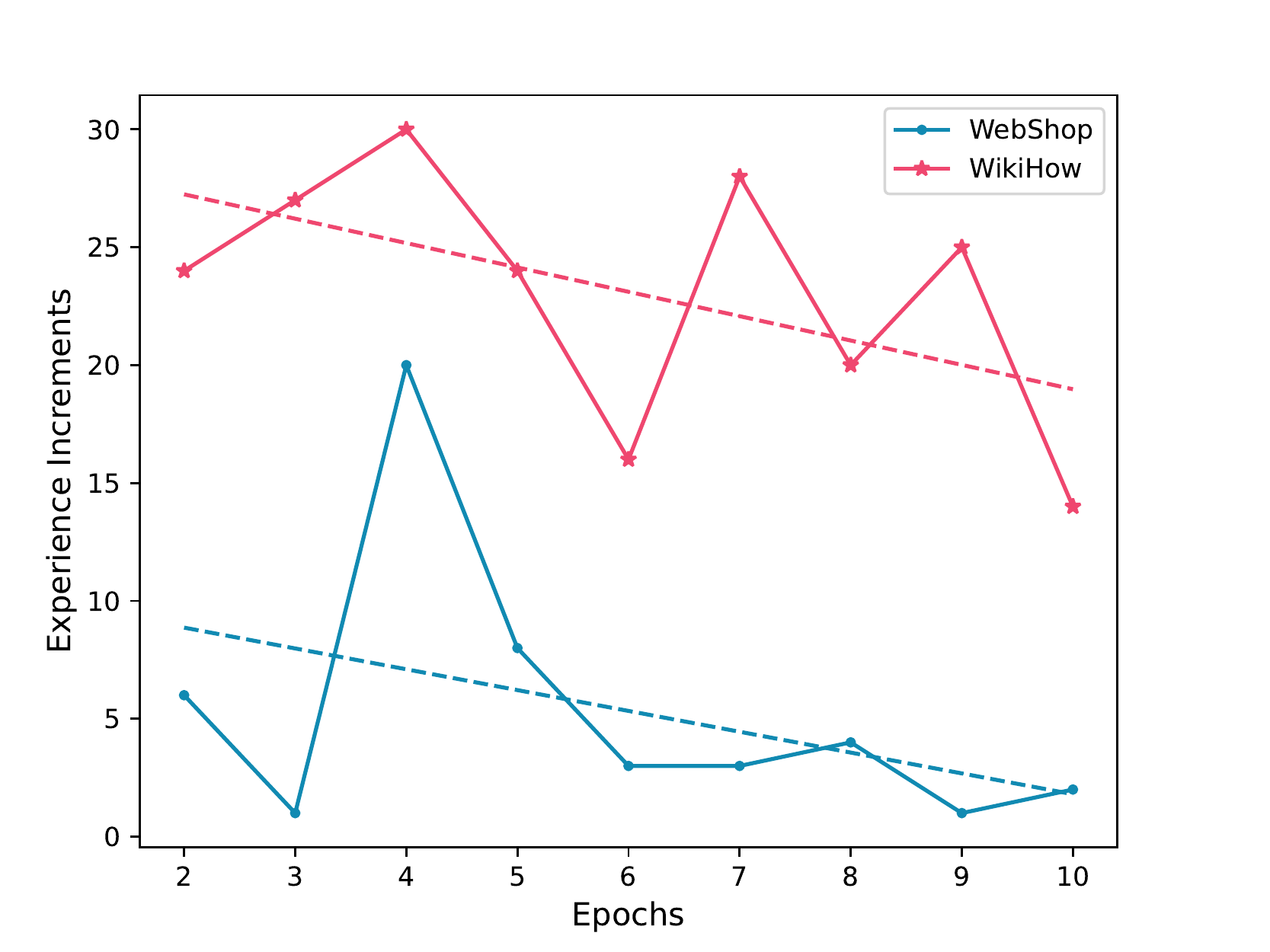}
		\end{center}
		\caption{Number of new experiences in each training epoch. The dashed
		lines are acquired by performing the least squares fit to the data
	points.}
		\label{fig:nb_incrm}
	\end{subfigure}
	\caption{Variation of the experience number in the training process}
	\label{fig:exp_var}
\end{figure}


We further conducted experiments to see how the capability of \Llmep{} evolves
during training. Owing to the limit of budgets, a subset of only 20 tasks is
sampled from the full test set. We visualize the performance on the subset of
\Llmep{} at epochs 1, 5, and 10. The performance at epoch 3, which is used for
the experiments in the main paper, is visualized as well. The visualization is
available in Figure~\ref{fig:cpbl_evlv}. It can be seen that the performance of
\Llmep{} improves during the training procedure. However, there seems to be a
saturation for the performance, which may be attributed to the limited number
of the active exemplars and training tasks. The saturation of the average
reward comes later than that of the success rate. This fact indicates that
\Llmep{} can still seize more rewards through training on several unsuccessful
tasks even the success rate has already saturated. In other words, the hard
tasks benefit more from the later phase of training than the easy tasks.
Besides, \Llmep{} reaches saturation on WebShop earlier than on WikiHow. To
give an explanation, the number of the experiences in the memory after each
training epoch is inspected. As shown in Figure~\ref{fig:exp_var}, there are
much fewer new experiences added into the memory in the later epochs for
WebShop than for WikiHow. The certain reason may be due to the specific
training set or some internal characteristics of the task domain, which will be
further investigated in the future work.

\section{$Q$ function fitting ability of \Llmep{}}

\begin{table}
	\centering
	\caption{Comparison of the average reward estimation of the full model and
	the Double Q-Learning model}
	\label{tab:dql_rst}
	\begin{tabular}{cccm{2cm}<{\centering}m{1.3cm}<{\centering}cc}
		\toprule[1.5pt]
		Task Set &	Setting &	\#Epochs	& Avg Reward Estimation &	Avg Training Reward &	Abs Error &	Relative Error \\
		\midrule
		\multirow{3}{*}{WebShop} &	Full Model &	3 &	0.86 &	0.84 &	\textbf{0.02} &	\textbf{2.38} \\
		&	+DoubleQL &	3 &	0.71 &	0.75 &	0.04 &	5.33 \\
		&	+DoubleQL &	6 &	0.69 &	0.77 &	0.08 &	10.39 \\
		\midrule
		\multirow{3}{*}{WikiHow} &	Full Model &	3 &	2.48 &	2.60 &	\textbf{0.12} &	\textbf{4.62} \\
		&	+DoubleQL &	3 &	2.47 &	2.90 &	0.43 &	14.83 \\
		&	+DoubleQL &	6 &	2.70 &	2.90 &	0.20 &	6.90 \\
		\bottomrule[1.5pt]
	\end{tabular}
\end{table}

Ablation study in the main paper has demonstrated that $n$-step bootstrapping
manages to improve precision of the learned $Q$ values in the memory. This
section will give further discussion about over-estimation of learned $Q$
values in the memory and whether the LLM can learn the certain $Q$ function
through in-context learning (ICL).

Double Q-Learning~\citep{HadoVanHasselt2010_DoubleQLearning} is usually
leveraged to ameliorate over-estimation for lookup-based Q-Learning.
Table~\ref{tab:dql_rst} shows the $Q$ value estimation results with Double
Q-Learning applied. Over-estimation does be suppressed, however, serious
under-estimation is introduced, and the estimation error fails to ameliorate.
This is explained by that Double Q-Learning iteratively updates two $Q$ value
lookups and requires more steps to converge to an accurate enough estimation.
In contrast, plain Q-Learning performs better in few-step circumstances.

As regards whether the LLM learns the certain $Q$ value function, predicted
values of LLM during the test phase on WebShop are inspected. The average
absolute error is 0.417. This fact indicates that the LLM does not really learn
the certain $Q$ function, as the reward in WebShop is always between 0 and 1.
Nevertheless, the LLM can still predict the appropriate actions. This is due to
the inessentiality of absolutely precise $Q$ value prediction during test.  It
is the relative relation between the values of candidate actions that is truly
important. Once LLM can distinguish the valuable actions from candidates, it
can take the right policy.

\section{Example of the exemplars}

\begin{figure}[h]
	\centering
	\includegraphics[width=\linewidth]{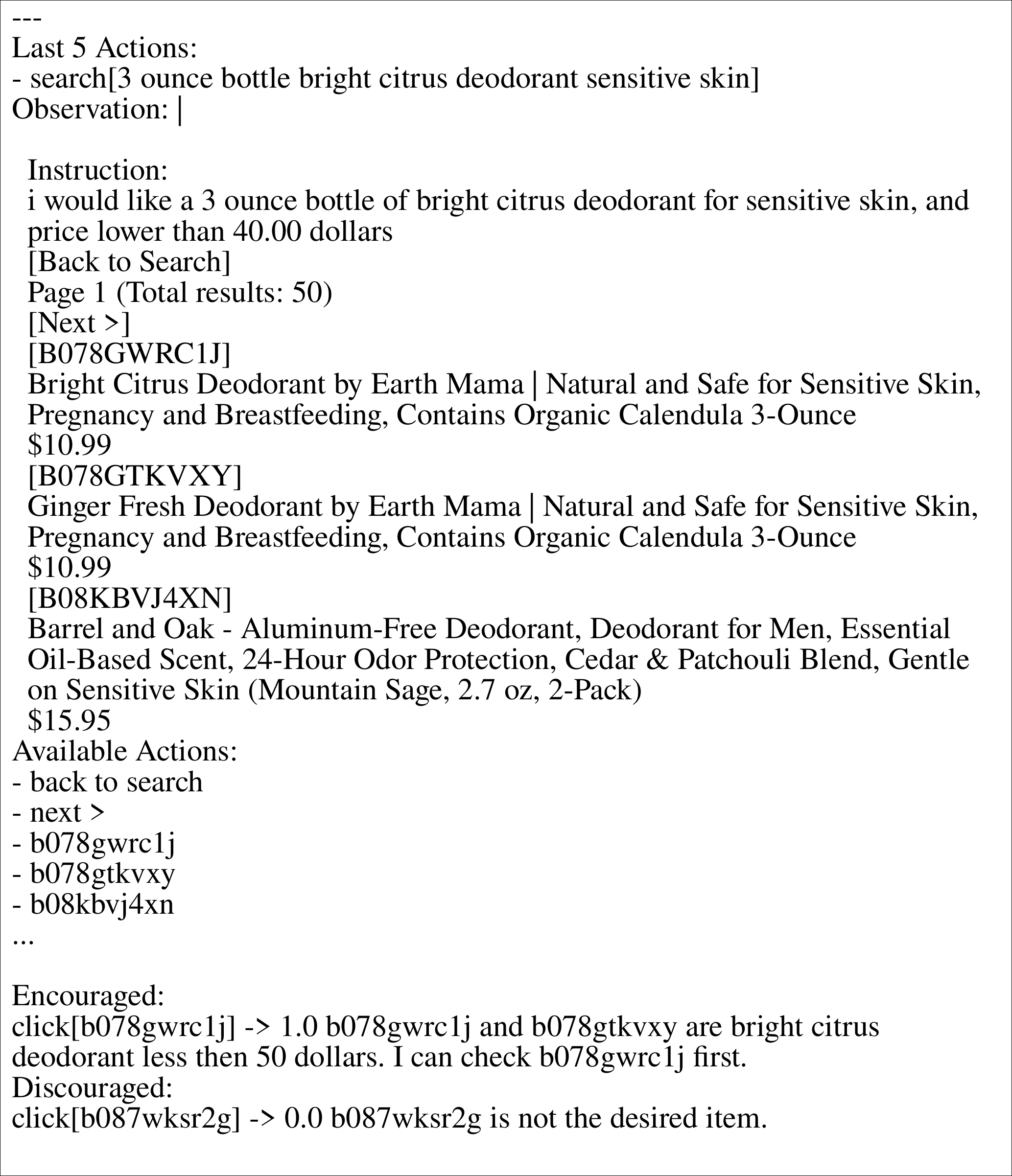}
	\caption{Exemplar for WebShop. YAML markups are adopted to avoid confusing
	the keywords like ``Observation:'' with the colon-ended titles in the page
representation.}
	\label{fig:wsh_exm}
\end{figure}

\begin{figure}[h]
	\centering
	\includegraphics[width=\linewidth]{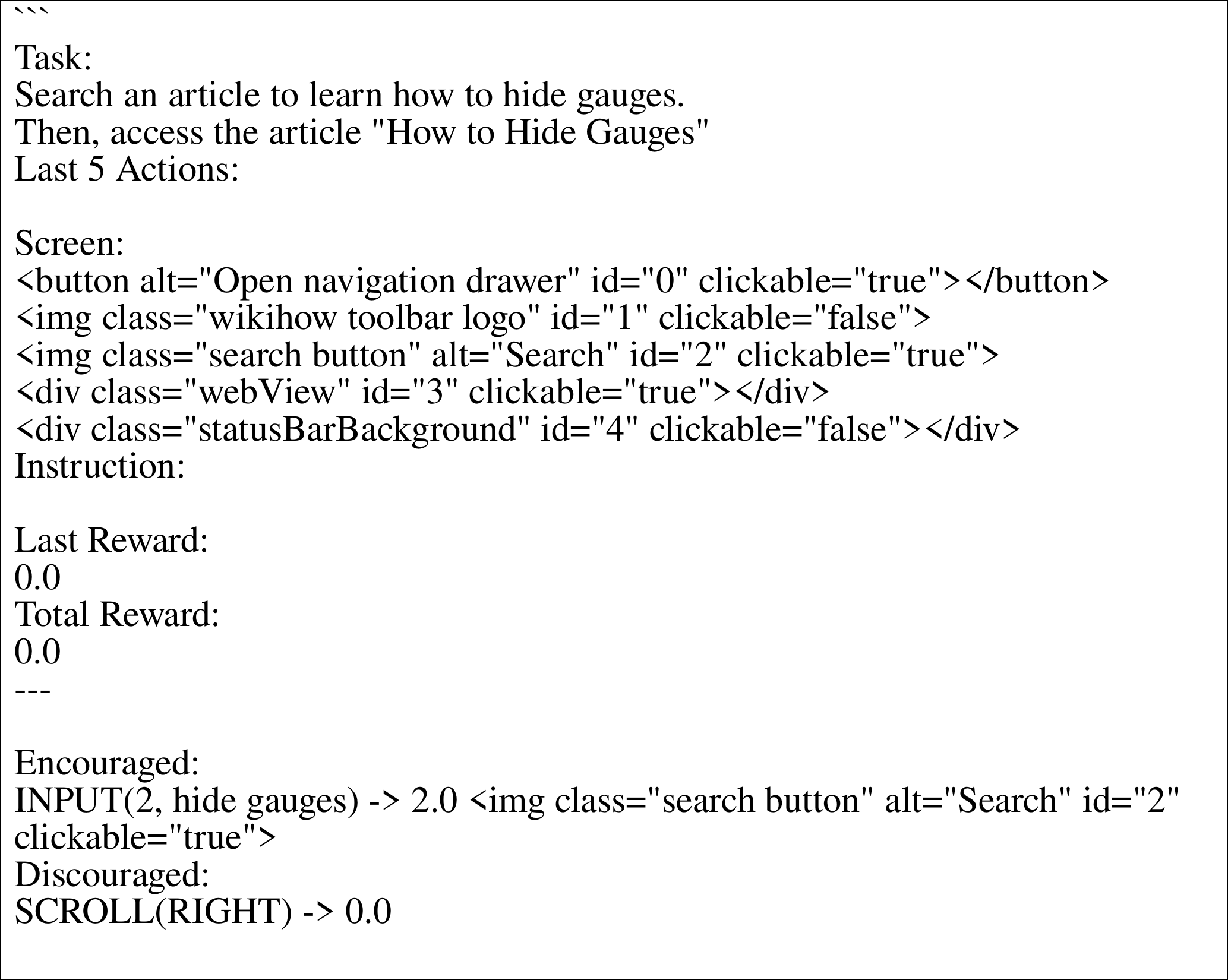}
	\caption{Exemplar for WikiHow}
	\label{fig:wkh_exm}
\end{figure}

An example of the input exemplar for WebShop and WikiHow is given in
Figure~\ref{fig:wsh_exm} and Figure~\ref{fig:wkh_exm}, respectively.

\section{Case study}

\begin{figure}[ht]
	\centering
	\includegraphics[width=\linewidth]{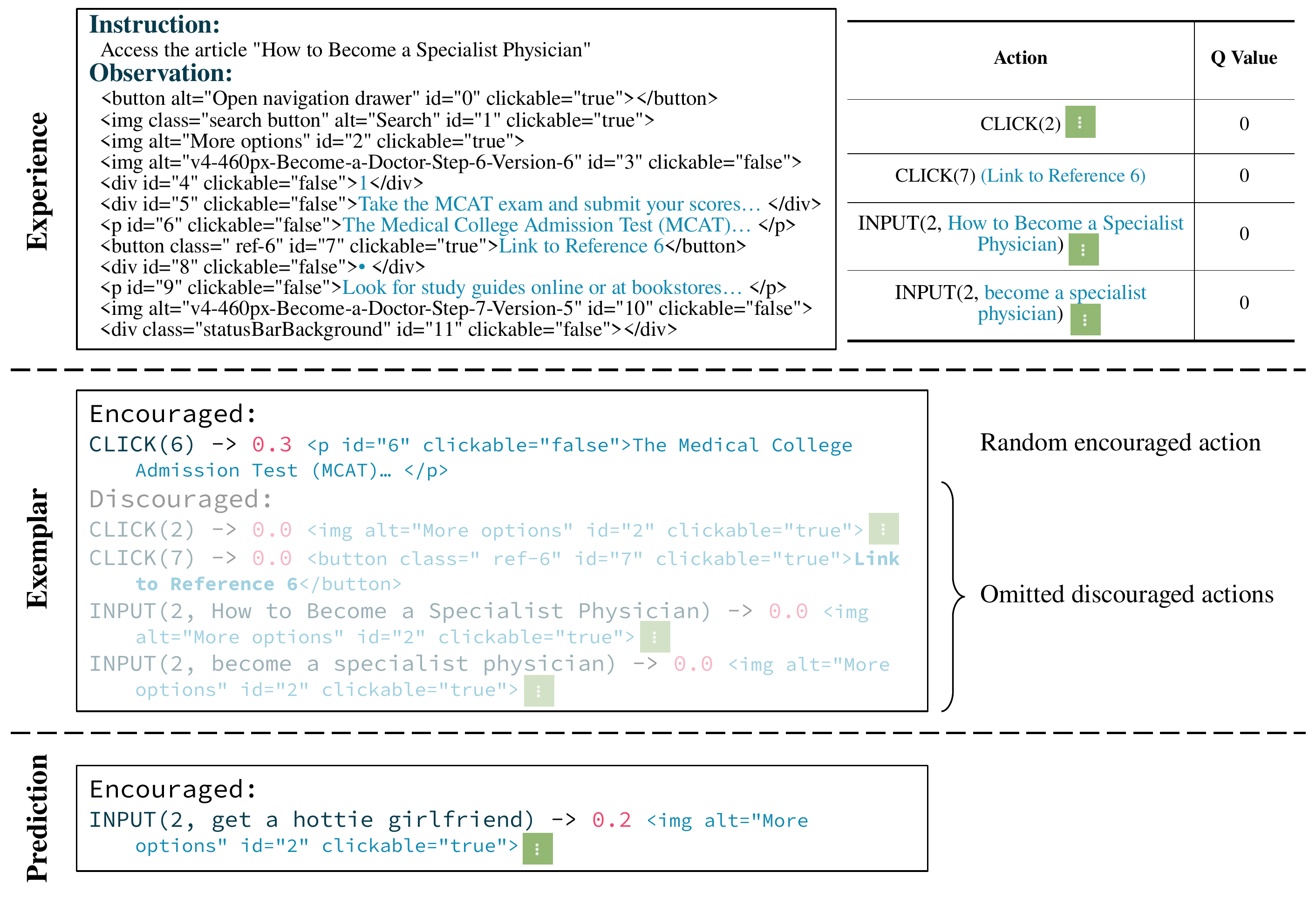}
	\caption{Case of the ablation study on the discouraged actions. As there
	are no valuable actions to encourage in the experience, a random action is
generated. When the discouraged actions with low value are omitted, the LLM may
repeat the failure with the same pattern.}
	\label{fig:dcr_act_c_std}
\end{figure}

\begin{figure}[ht]
	\centering
	\includegraphics[width=\linewidth]{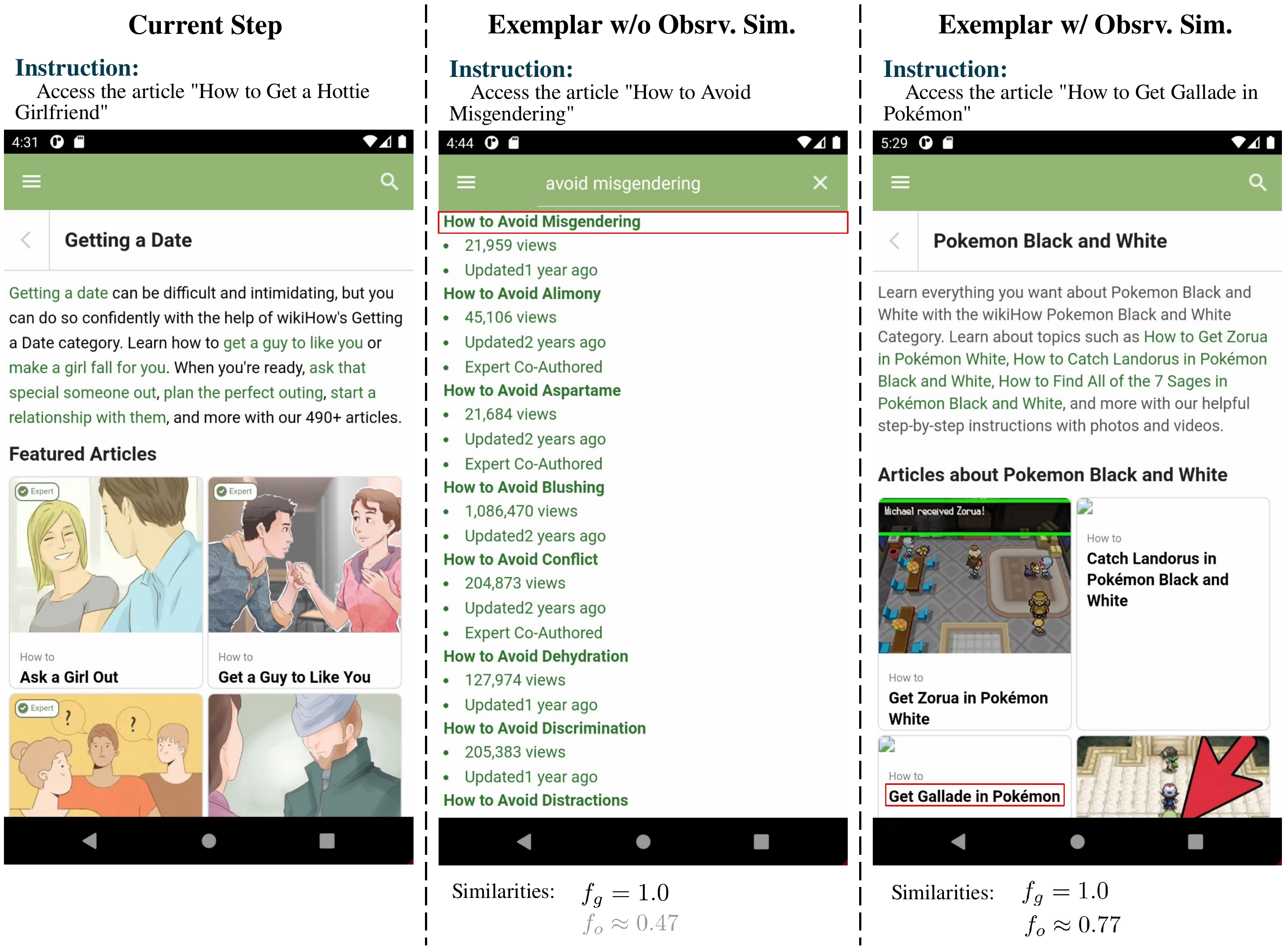}
	\caption{Case of the ablation study on the similarity function.  Encouraged
	actions recorded in the experiences are marked by red rectangles.}
	\label{fig:sim_fct_c_std}
\end{figure}

Figure~\ref{fig:dcr_act_c_std} gives a case from the ablation study on
necessity of the discouraged actions. If the discouraged actions are omitted in
the action advice from an experience without encouraged actions, the LLM will
have no ability to avoid failures of the same pattern.

A case from the ablation study on the similarity function on WikiHow task set
is depicted in Figure~\ref{fig:sim_fct_c_std}. Once the observation similarity
$f_o$ is omitted, the agent will retrieve experience only according to the
instruction and cannot adjust the selection in accordance with the particular
observation. This will cause improper experience retrieval and lead to poorer
performance.

\end{document}